\documentclass{bmvc2k}
\usepackage{amsmath,amsfonts}
\usepackage{amssymb,amsopn}
\usepackage{bm} 

\usepackage{amssymb}
\usepackage{amsmath,amsfonts}
\usepackage{amsthm,amsopn}

\usepackage{bm} 

\newcommand{\ProbOpr}[1]{\mathbb{#1}}

\newcommand{\expect}[2]{%
\ifthenelse{\equal{#2}{}}{\ProbOpr{E}_{#1}}
{\ifthenelse{\equal{#1}{}}{\ProbOpr{E}\left[#2\right]}{\ProbOpr{E}_{#1}\left[#2\right]}}} 
\newcommand{\var}[2]{%
\ifthenelse{\equal{#2}{}}{\ProbOpr{VAR}_{#1}}
{\ifthenelse{\equal{#1}{}}{\ProbOpr{VAR}\left[#2\right]}{\ProbOpr{VAR}_{#1}\left[#2\right]}}} 

\DeclareMathOperator{\argmax}{arg\,max}

\newcommand{\eat}[1]{}
\usepackage{comment}
\usepackage{floatrow}
\usepackage{multirow}
\newfloatcommand{capbtabbox}{table}[][\FBwidth]

\title{An Empirical Study on Leveraging Scene Graphs for Visual Question Answering}

\addauthor{Cheng Zhang}{zhang.7804@osu.edu}{1}
\addauthor{Wei-Lun Chao}{weilunchao760414@gmail.com}{12}
\addauthor{Dong Xuan}{xuan.3@osu.edu}{1}

\addinstitution{
 The Ohio State University \\
 Columbus, Ohio, USA
}
\addinstitution{
 Cornell University \\
 Ithaca, New York, USA
}

\runninghead{Zhang, Chao, Xuan}{Leveraging Scene Graphs for Visual QA}

\renewcommand{\paragraph}[1]{\vspace{1ex}\noindent\textbf{#1}}
\def\ie{\emph{i.e}\bmvaOneDot}
\def\eg{\emph{e.g}\bmvaOneDot}

\def\etal{\emph{et al}\bmvaOneDot}
\newcommand{\method}[1]{\textsc{#1}}
\newcommand{\uGN}{\method{u-GN}} 
\newcommand{\fGN}{\method{f-GN}} 
\newcommand{\cmmnt}[1]{}
\begin{document}
\maketitle
\begin{abstract}
Visual question answering (Visual QA) has attracted significant attention these years. While a variety of algorithms have been proposed, most of them are built upon different combinations of image and language features as well as multi-modal attention and fusion. In this paper, we investigate an alternative approach inspired by conventional QA systems that operate on knowledge graphs. Specifically, we investigate the use of scene graphs derived from images for Visual QA: an image is abstractly represented by a graph with nodes corresponding to object entities and edges to object relationships. We adapt the recently proposed graph network (GN) to encode the scene graph and perform structured reasoning according to the input question. Our empirical studies demonstrate that scene graphs can already capture essential information of images and graph networks have the potential to outperform state-of-the-art Visual QA algorithms but with a much cleaner architecture. By analyzing the features generated by GNs we can further interpret the reasoning process, suggesting a promising direction towards explainable Visual QA.

\end{abstract}
\section{Introduction}\label{sec:intro}

Scene understanding and reasoning has long been a core task that the computer vision community strives to advance. In recent years, we have witnessed significant improvement in many representative sub-tasks such as object recognition and detection, in which the machines' performance is on par or even surpasses humans'~\cite{he2015delving,he2016deep}, motivating the community to move toward higher-level sub-tasks such as visual captioning~\cite{anderson2018bottom,mao2015deep,xu2015show} and visual question answering (Visual QA)~\cite{anderson2018bottom,antol2015vqa,yang2016stacked,zhu2016visual7w}.

\begin{figure}
    \caption{\small Scene graphs for Visual QA. We show the image, the human annotated graph~\cite{krishna2017visual}, and the machine generated graph~\cite{zellers2018neural}. The answer can clearly be reasoned from the scene graph.}
    \vskip -5pt
    \centerline{\includegraphics[width=1\linewidth]{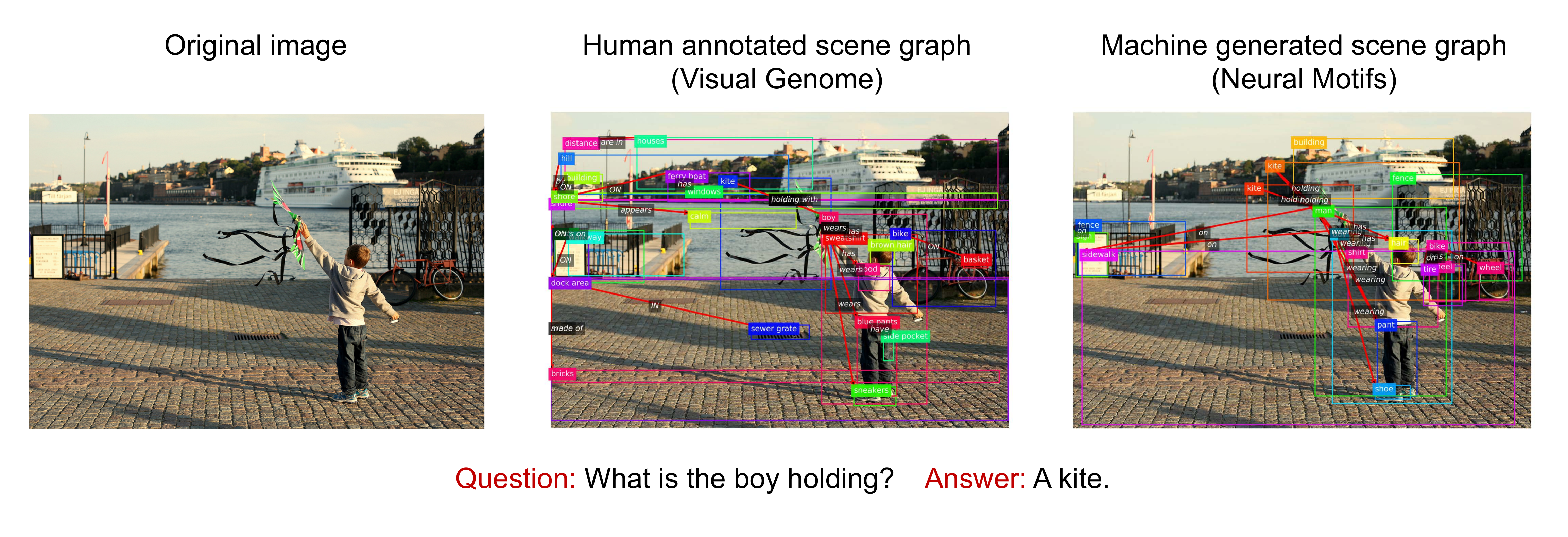}}
    \vskip -20pt
    \label{f1}
\end{figure}

One key factor to the recent successes is the use of neural networks~\cite{goodfellow2016deep}, especially the convolutional neural network (CNN)~\cite{lecun1995convolutional} which captures the characteristics of human vision systems and the presence of objects in a scene. An object usually occupies nearby pixels and its location in the image does not change its appearance much, making CNNs, together with region proposals~\cite{uijlings2013selective}, suitable models for object recognition and detection. Indeed, when CNN~\cite{krizhevsky2012imagenet} and R-CNN~\cite{girshick2014rich,ren2015faster} were introduced, we saw a leap in benchmarked performance.

Such a leap, however, has yet to be seen in high-level tasks. Taking Visual QA~\cite{antol2015vqa,zhu2016visual7w,goyal2017making,krishna2017visual} as an example, which takes an image and a question as inputs and outputs the answer. In the past four years there are over a hundred of publications, but the improvement of the most advanced model over a simple multi-layer perceptron (MLP) baseline is merely $5\sim7\%$~\cite{anderson2018bottom,hu2018learning,yu2018beyond}. Besides, the improvement is mostly built upon deep image and language features~\cite{he2016deep,hochreiter1997long,mikolov2013distributed}, multi-modal fusion~\cite{ben2017mutan,fukui2016multimodal}, and attention~\cite{anderson2018bottom,yang2016stacked}, not a fundamental breakthrough like CNNs in object recognition.

Interestingly, if we take the visual input away from Visual QA, there has been a long history of development in question answering (QA), especially on leveraging the knowledge graphs (or bases)~\cite{bordes2014question,berant2013semantic,lukovnikov2017neural,yao2014information,yih2015semantic}. The basic idea is to represent the knowledge via entities and their relationships and then query the structured knowledge during testing time. In the vision community, we have also seen attempts to construct the so-called scene graph~\cite{johnson2015image,krishna2017visual} that can describe a visual scene in a similar way to a knowledge graph~\cite{li2017scene,li2018factorizable,wang2018scene,xu2017scene,yang2018graph,zellers2018neural}. Nevertheless, we have not seen a notable improvement or comprehensive analysis in exploiting the structured scene graph for Visual QA, despite the fact that several recent works have started to incorporate it~\cite{ben2019block,cadene2019murel,li2019relation,liang2019rethinking,shi2018explainable,yang2018multi}.

There are multiple possible reasons. Firstly, we may not yet have algorithms to construct high-quality scene graphs. Secondly, we may not yet have algorithms to effectively leverage scene graphs. 
Thirdly, perhaps we do not explicitly need scene graphs for Visual QA: either they do not offer useful information or existing algorithms have implicitly exploited them.

In this paper we aim to investigate these possible reasons. Specifically, we take advantage of the recently published graph network (GN)~\cite{battaglia2018relational}, which offers a flexible architecture to encode nodes (\eg, \emph{object entities} and \emph{attributes}), edges (\eg, \emph{object relationships}), and global graph properties as well as perform (iterative) structured computations among them. By treating the question (and image) features as the input global graph properties and the answer as the output global properties, GNs can be directly applied to incorporate scene graphs and be learned to optimize the performance of Visual QA.

We conduct comprehensive empirical studies of GNs on the Visual Genome dataset~\cite{krishna2017visual,chao2017being}, which provides human annotated scene graphs. Recent algorithms on automatic scene graph generation from images~\cite{li2017scene,li2018factorizable,wang2018scene,xu2017scene,yang2018graph,zellers2018neural} have also focused on Visual Genome, allowing us to evaluate these algorithms using Visual QA as a \cmmnt{down-stream}down-stream task (see Fig.~\ref{f1}). Our experiments demonstrate that human annotated or even automatically generated scene graphs have already captured essential information for Visual QA. Moreover, applying GNs without complicated attention and fusion mechanisms shows promising results but with a much cleaner architecture. By analyzing the GN features along the scene graph we can further interpret the reasoning process, making graph networks suitable models to leverage scene graphs for Visual QA tasks. 
\section{Related Work}\label{sec:relatedWork}
\vspace{-3mm}
\paragraph{Visual question answering (Visual QA).}
Visual QA requires comprehending and reasoning with visual and textual information. Existing algorithms mostly adopt the pipeline that first extracts image and question features~\cite{jabri2016revisiting}, followed by multi-modal fusion~\cite{ben2017mutan,fukui2016multimodal} and attention~\cite{anderson2018bottom,lu2016hierarchical,yang2016stacked} to obtain multi-modal features to infer the answer. While achieving promising results, it remains unclear if the models have been equipped with reasoning abilities or solely rely on exploiting dataset biases. Indeed, the performance gain between a simple MLP baseline and a complicated attention model is merely within $5\sim 7\%$~\cite{hu2018learning}. The improvement brought by newly proposed models usually lies within $0.5\%$~\cite{ben2017mutan,mun2018learning,anderson2018bottom}.

Some new paradigms thus aim to better understand the intrinsic behavior of models~\cite{kafle2017analysis}. One direction is to leverage object relationships for better scene understanding~\cite{johnson2017clevr,santoro2017simple,shi2018explainable,liang2019rethinking,li2019relation}. The other aims for interpretable Visual QA via neuro-symbolic learning~\cite{mao2018neuro,yi2018neural,vedantam2019probabilistic}, developing symbolic functional programs to test machines' reasoning capability in Visual QA. Nevertheless, many of \cmmnt{those}these works mainly experiment on synthetic data.

\paragraph{QA with knowledge bases.}
In conventional QA systems without visual inputs, exploiting knowledge bases (KBs)~\cite{fader2014open,bollacker2008freebase,hoffart2011yago2} to store complex structured and unstructured information so as to support combinatorial reasoning has been widely investigated~\cite{fader2014open,bordes2014question,berant2013semantic,lukovnikov2017neural,yao2014information,yih2015semantic}. There are \cmmnt{in general}generally two kinds of KBs~\cite{fader2014open}: curated and extracted. Curated KBs, such as Freebase~\cite{bollacker2008freebase} and YAGO2~\cite{hoffart2011yago2}, extract $\mathsf{<entity-relationship-entity>}$ triples from knowledge sources like Wikipedia and WordNet~\cite{miller1995wordnet}. Extracted KBs~\cite{banko2007open,carlson2010toward}, on the other hand, extract knowledge in the form of natural language from millions of web pages. 

The scene graph~\cite{johnson2015image,krishna2017visual} examined in our work could be seen as a form of curated KBs that extracts triplets from images. Specifically, an image will be abstractly represented by a set of object entities and their relationships. Some Visual QA works indeed extract scene graphs followed by applying algorithms designed for reasoning on curated KBs~\cite{krishnamurthy2013jointly, wang2018fvqa}. In our work we also exploit scene graphs, but apply the recently published graph networks (GNs)~\cite{battaglia2018relational} that can 
easily incorporate advanced deep features for entity and relationship representations and allow end-to-end training to optimize overall performance.

\paragraph{Scene graphs.}
Lately, generating scene graphs from images has achieved notable progress\cmmnt{lately} \cite{xu2017scene,zellers2018neural,li2017scene,newell2017pixels,yang2018graph,krishna2018referring}. As an abstraction of images, scene graphs have been shown to improve image retrieval~\cite{johnson2015image}, generation~\cite{johnson2018image}, captioning~\cite{gupta2008beyond,anderson2016spice,yao2018exploring}, video understanding~\cite{ma2018attend,vicol2018moviegraphs}, and human object interaction~\cite{kato2018compositional}.

Recent works on Visual QA~\cite{ben2019block,cadene2019murel,li2019relation,liang2019rethinking,shi2018explainable,yang2018multi} and visual reasoning~\cite{shi2018explainable,johnson2017inferring} have also begun to \cmmnt{also start to }exploit scene graphs. However, they usually integrate multiple techniques (\eg, scene graph generation, attention, multi-modal fusion) into one hybrid model to obtain state-of-the-art performance. It is thus hard to \cmmnt{tell}conclude if scene graphs truly contribute to the improvement. Indeed, in an early study~\cite{wu2017visual} of the Visual Genome dataset, only $40\%$ of the questions could \cmmnt{exactly be answered}be answered exactly with the human annotated scene graphs, assuming that the reasoning on the graphs is perfect\footnote{In the study~\cite{wu2017visual}, a question is considered answerable by the scene graph if its answer exactly matches any node or relationship names or their combinations from the scene graph.}. This ratio is surprisingly low considering the detailed knowledge encoded by scene graphs and the existing Visual QA models' performance (\eg, $>50\%$~\cite{hu2018learning}). 

We therefore perform a systematic study on leveraging scene graphs without applying specifically designed attention and fusion mechanisms. The results \cmmnt{thus can}can then faithfully indicate the performance gain brought by performing structured reasoning on scene graphs.
\section{Leveraging Scene Graphs for Visual QA}
\label{sec:approach}

\begin{figure*}
    \centerline{\includegraphics[width=1\linewidth]{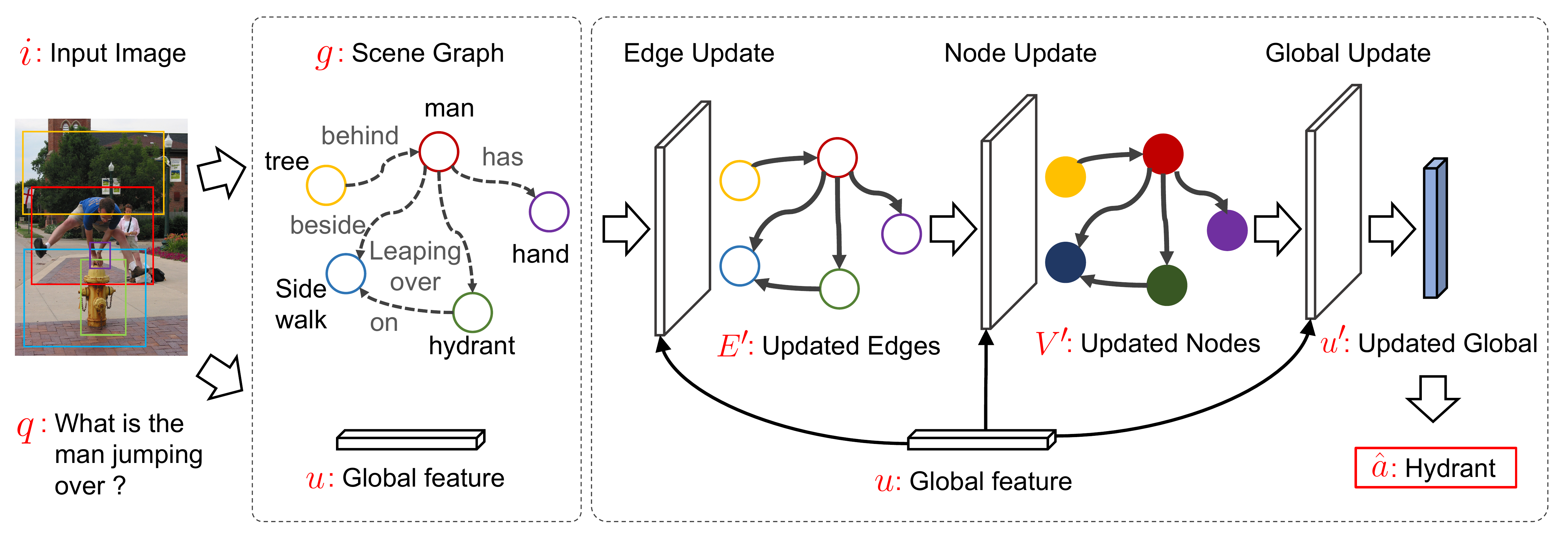}}
    \vskip -10pt 
    \caption{\small The GN-based Visual QA framework. The unfilled (filled) nodes, dashed (solid) edges, and unfilled (filled) cuboids denote the un-updated (updated) features. The updated global features (filled cuboid) are used to predict the answer.}
    \label{f2}
\end{figure*}

In this section we describe the graph network (GN) architecture~\cite{battaglia2018relational} and how we apply it to reason from a scene graph according to the input image and question for Visual QA. Fig.~\ref{f2} gives an illustration of the GN-based Visual QA framework. In what follows, we first define the Visual QA task and the scene graph, and then introduce a general Visual QA framework.

\subsection{Problem definitions}
\label{sec:approach_def}
A Visual QA model takes an image $i$ and a related question $q$ as inputs, and needs to output the correct answer $a$. In this work, we consider the multiple-choice Visual QA setting~\cite{antol2015vqa,zhu2016visual7w,chao2017being}, in which the model needs to pick the correct answer $a$ from a set of $K$ candidate answers $\mathcal{A}=\{a, d_1, \cdots, d_{K-1}\}$. $d_k$ is called a negative answer or decoy. Nevertheless, our model can be easily extended into the open-ended setting with the answer embedding strategy~\cite{hu2018learning}.  

We explicitly consider encoding the image $i$ via a directed scene graph $g=(V, E)$~\cite{johnson2015image}. The node set $V=\{v_n\}_{n=1:N}$ contains $N$ objects of the image, where each $v_n$ records an object's properties such as its name (\eg, a car or a person), attributes (\eg, colors, materials), location, and size. The edge set $E=\{(e_m, s_m, o_m)\}_{m=1:M}$ contains $M$ pairwise relationships between nodes, where $e_m$ encodes the relationship name (\eg, on the top of); $s_m, o_m$ are the indices of the subject and object nodes, respectively. For now let us assume that $g$ is given for every $i$. We note that $N$ and $M$ may vary among different images.

A general Visual QA model thus can be formulated as follows~\cite{jabri2016revisiting,chao2017being},
\begin{align}
\hat{a}  = \argmax_{c\in\mathcal{A}} h_\theta(c, i, q, g), \label{eq:VQA}
\end{align}
where $c\in\mathcal{A}$ is a candidate answer and $h_\theta$ is a learnable scoring function on how likely $c$ is the correct answer of $(i, q, g)$. 

In Visual QA without the scene graph $g$, $h_\theta(c,i,q)$ can be modeled by an MLP with the concatenated features of $(c,i,q)$ as the input. For example,~\cite{jabri2016revisiting,chao2017being} represents $c$ and $q$ via the average word vectors and $i$ via CNN features. Alternatively, one can factorize $h_\theta(c,i,q)$ by $\gamma_\theta(\alpha_\theta(i, q), \beta_\theta(c))$, where $\alpha_\theta(i, q)$ is the joint image-question embedding and $\beta_\theta(c)$ is the answer embedding~\cite{hu2018learning}. $\gamma_\theta$ measures the compatibility of the two embeddings, \eg, by inner products.
We denote the former as unfactorized models and the later as factorized models.

In Visual QA with scene graphs, we must define $h_\theta(c,i,q,g)$ so that it can take the nodes $V$ and edges $E$ into account as well as maintain the permutation invariant characteristic of a graph\footnote{That is, even we permute the indices of the nodes or edges, the output of $h_\theta$ should be the same.}. Moreover, $h_\theta(c,i,q,g)$ must be able to reason from the graph $g$ according to $(c, i, q)$. In the next subsection we introduce the graph network (GN), which provides a flexible computational architecture to fulfill the requirements above.

\subsection{Graph networks (GNs)} \label{sec:approach_GN}
The graph network (GN) proposed in~\cite{battaglia2018relational} defines a computational block (module) that performs \textit{graph-to-graph} mapping. That is, it takes a graph $g = (u, V, E)$ as the input and outputs an updated graph $g' = (u', V', E')$, which has the same graphical structure but different encoded information (\ie, $v_n$ and $e_m$ will be updated). \emph{The components $u$ and $u'$ encode certain global properties (features) of the graph.} For example, in Visual QA, $u$ can be used to encode the image $i$ and question $q$, while $u'$ can be used to predict the final answer $a$.

There are various ways to define the updating procedure, as long as it maintains permutation invariant with respect to the graph. Here we describe a procedure that updates edges, nodes, and global features in order. We will then discuss how this updating procedure is particularly suitable for Visual QA with scene graphs.

\paragraph{Edge updates.} The edge updating function $f^e_\theta$ is performed per edge $m$ according to its features (encoded information) $e_m$, the features of the subject and object nodes $v_{s_{m}}$ and ${v}_{o_{m}}$, and the global features $u$,
\begin{equation}\label{edge-1}
    e'_m= f^{e}_\theta(e_m,v_{s_{m}}, {v}_{o_{m}}, u).
\end{equation}

\paragraph{Node updates.} The subsequent node updates first aggregate the information from incoming edges of each node $n$,
\begin{equation}\label{node-1}
    \overline{e'_n}= \phi^{e\rightarrow v}(E'_n),
\end{equation}
where $\phi^{e\rightarrow v}$ is an aggregation function and $E'_n=\{(e'_m,s_m,o_m)\}_{o_{m}=n}$
is the set of incoming edges.
We then apply the node updating function $f^v_\theta$ to each node $n$,
\begin{equation}\label{node-2}
    v'_{n}= f^{v}_\theta(v_{n},\overline{e'_n}, u).
\end{equation}

\paragraph{Global updates.} Finally, the global updating function $f^{u}_\theta$ is applied to update the global features. It begins with aggregating the edge and node features,
\begin{align}\label{u-2}
    \overline{e'}= \phi^{e\rightarrow u}(E'),\\
    \overline{v'}= \phi^{v\rightarrow u}(V'),
\end{align}
where $E'=\{(e'_m,s_m,o_m)\}_{m=1:M}$ and $V'=\{v'_n\}_{n=1:N}$. The updated $u'$ is then computed by
\begin{equation}\label{u-1}
    u'= f^{u}_\theta(\overline{e'}, \overline{v'}, u).
\end{equation}

\smallskip 

In our studies, we assume that $u, v_n, e_m$ (and $u', v'_n, e'_m$) each can be represented by a real vector. We make the following choices of aggregation and updating functions. The aggregation functions $\phi^{e\rightarrow v}, \phi^{e\rightarrow u}, \phi^{v\rightarrow u}$ should be able to take various number of inputs and must be permutation invariant with respect to the indices of their inputs. Representative options are element-wise average, sum, and max operations and we use element-wise average in the paper. The resulting $\overline{e'_n}$, $\overline{e'}$, and $\overline{v'}$ are thus real vectors.

The updating functions $f^e_\theta$, $f^v_\theta$, and $f^u_\theta$ can be any learnable modules such as MLPs and CNNs. In this paper, we apply MLPs with the concatenated features as the input. We note that, all the updating functions are optimized simultaneously; all the edges and nodes share the same updating functions $f^e_\theta$ and $f^v_\theta$. More importantly, all the updating functions can be shared across different graphs even if they have different structures.

The resulting graph $g' = (u', V', E')$ then can be used to perform inference for tasks like Visual QA, or serve as the input to a subsequent GN block.

\subsection{GN-based Visual QA with scene graphs}
\label{sec:GN_VQA}
We apply GNs to Visual QA with scene graphs for multiple reasons. First, GNs explicitly consider the graph structures in its computations and can share the learned functions across different graphs\footnote{Different images may have different scene graphs with varied numbers of nodes and edges.}. Second, by encoding the question features (as well as the image and candidate answer features) into the global graph features $u$, GNs directly supports reasoning on graphs with appropriate choices of updating functions and procedures. Finally, GNs can be easily incorporated into Eq.~(\ref{eq:VQA}) and learned to optimize the overall Visual QA performance. In the following we give more details.

\paragraph{Features.}
We encode $q$ and $c$ (a candidate answer) with averaged word vectors~\cite{mikolov2013distributed} and $i$ with CNN features~\cite{he2016deep}, following~\cite{jabri2016revisiting,chao2017being,hu2018learning}. We obtain the scene graph of each image either via human annotation~\cite{krishna2017visual} or via automatic scene graph generation~\cite{zellers2018neural}. Ideally, every node in a graph will be provided with a node name (\eg, car) and a set of attributes (\eg, red, hatchback) and we represent each of them by the averaged word vectors. We then concatenate them to be the node features $v_n$. For edges that are provided with the relationship name (\eg, on the top of), we again encode each $e_m$ by the average word vectors. We note that more advanced visual and natural language features~\cite{devlin2018bert} can be applied to further improve the performance.

\paragraph{Learning unfactorized Visual QA models.}
We model the scoring function $f_\theta(c,i,q,g)$ in Eq.~(\ref{eq:VQA}) as below,
\begin{align}\label{eq:u-GN}
    f_\theta(c,i,q,g) = u' & \hspace{10pt}\text{s.t.}\hspace{10pt}(u', V', E') = GN(u = [c, i, q], V, E),
\end{align}
where $[\cdot]$ is a concatenation operation. That is, the output of the GN global updating function $f_\theta^u$ is a scalar indicating whether $c$ is the correct answer of the image-question pair. We learn the parameters of the GN block to optimize the binary classification accuracy, following~\cite{jabri2016revisiting}. We note that, even if only $u'$ is being used to predict the answer, all the three updating functions will be learned jointly according to the updating procedures in Sect.~\ref{sec:approach_GN}.

\paragraph{Learning factorized Visual QA models.}
We can further factorize $f_\theta(c,i,q,g)$ in Eq.~(\ref{eq:VQA}) by $\gamma_\theta(\alpha_\theta(i,q,g), \beta_\theta(c))$~\cite{hu2018learning}.  Specifically, we model $\alpha_\theta(i,q,g)$ by a GN block, $\beta_\theta(c)$ by an MLP, and $\gamma_\theta$ by another MLP following~\cite{conneau2017supervised},

\begin{align}\label{eq:f-GN}
    & \alpha_\theta(i, q, g) = u' \hspace{10pt}\text{s.t.}\hspace{10pt} (u', V', E') = GN(u = [i, q], V, E),\\
    & \beta_\theta(c) = c' \hspace{10pt}\text{s.t.}\hspace{10pt} c'= MLP(c), \nonumber\\ 
    & f_\theta(c,i,q,g) = \gamma_\theta(u', c') = MLP([c', u', |c' - u'|, c' * u']), \nonumber
\end{align}
where $*$ indicates element-wise product. The factorized model offers faster training and inference and can be used for the open-ended setting directly: the GN block will be computed just once for an image-question pair no matter how many candidate answers are considered.
\vspace{-3mm}
\section{Experiments}\label{sec:exp}
\subsection{Setup}\label{sec:setup}
\paragraph{Visual QA dataset.} 
We conduct experiments on the \textit{Visual Genome (VG)}~\cite{krishna2017visual} dataset. VG contains 101,174 images annotated with in total 1,445,322 $(i, q, a)$ triplets. Chao \etal~\cite{chao2017being} further augment each triplet with 6 auto-generated decoys (incorrect answers) and split the data into 727K/283K/433K triplets for training/validation/testing, named \emph{qaVG}. Following~\cite{chao2017being,hu2018learning}, we evaluate the accuracy of picking the correct answer from 7 candidates.

\paragraph{Scene graphs.} 
We evaluate both human annotated and machine generated scene graphs. VG directly provides human-annotated graphs~\cite{krishna2017visual}, and we obtain machine generated ones by the start-of-the-art Neural Motifs (NM)~\cite{zellers2018neural}. Since the released NM model is trained from a different data split of \emph{qaVG}, we re-train the model from scratch using the training images of \emph{qaVG}. One key difference between the ground-truth and MotifsNet graphs is that the first ones provide both object names and attributes for nodes, while the later only provide names.

\paragraph{Baseline methods.} 
We compare with the unfactorized MLP models~\cite{jabri2016revisiting,chao2017being} and the factorized fPMC(MLP) and fPMC(SAN)~\cite{hu2018learning} models. SAN stands for stacked attention networks~\cite{yang2016stacked}. We note that, while the MLP model is extremely simple, it is only outperformed by fPMC(SAN) with layers of attentions by $5\%$.

\paragraph{Variants of our models.}
We denote the unfactorized GN model in Eq.~(\ref{eq:u-GN}) as \textbf{\uGN} and the factorized GN model in Eq.~(\ref{eq:f-GN}) as \textbf{\fGN}. We compare different combinations of input global features $u$ (\ie, $[c,i,q], [c,q]$ for \uGN{} and $[i,q], [q]$ for \fGN{}) with or without image features. We also compare different node features, with or without attributes. Finally, we consider removing edges (\ie, all nodes are isolated) or even removing the whole graph (\ie, Visual QA using the global features only).

\paragraph{Implementation details.}
We have briefly described the features in Sect.~\ref{sec:GN_VQA}. We apply $300$-dimensional word vectors~\cite{mikolov2013distributed} to represent $q$, $c$ (a candidate answer), node names, node attributes, and edge (relationship) names. If a node has multiple attributes, we again average their features. We apply ResNet-152~\cite{he2016deep} to extract the 2,048-dimensional image features. \emph{We perform $\ell_2$ normalization to each features before inputting them to the updating functions.}

For the GN block, we implement each updating function by a one-hidden-layer MLP whose architecture is a fully-connected (FC) layer followed by batch normalization~\cite{ioffe2015batch}, ReLU, Dropout (0.5)~\cite{srivastava2014dropout}, and another FC layer. The hidden layer is 8,192-dimensional. For $f^v_\theta$ and $f^e_\theta$, we keep the output size the same as the input. For $f^u_\theta$, \uGN{} has a 1-dimensional output while \fGN{} has a 300-dimensional output. For  \fGN{}, $\beta_\theta$ and $\gamma_\theta$ are implemented by the same MLP architecture as above with a 8,192-dimensional hidden layer. $\beta_\theta$ has a 300-dimensional output while $\gamma_\theta$ has a 1-dimensional output.

We learn both \uGN{} and \fGN{} using the binary classification objective (\ie, if $c$ is the correct answer or not). We use the Adam optimizer~\cite{kingma2014adam} with an initial learning rate $10^{-3}$ and a batch size 100. We divide the learning rate by $10$ whenever the validation accuracy decreases. We train for at most $30$ epochs and pick the best model via the validation accuracy.

\subsection{Main results}\label{sec:mainRes}
We summarize the main results in Table~\ref{tab:my_label}, in which we compare our \uGN{} and \fGN{} models to existing models on multiple-choice Visual QA. Both \uGN~(NM graphs) and \fGN~(VG graphs) outperform existing unfactorized methods, while \fGN~(VG graphs) is outperformed by fPMC(SAN$\star$) by $1\%$. We note that fPMC(SAN) uses LSTM~\cite{hochreiter1997long} to encode questions and fPMC(SAN$\star$) additionally uses it to encode answers. We thus expect that our GN models to have improved performance with better language features.

We then analyze different variants of our models. Comparing \uGN~(No graphs) to \uGN~(VG graphs) and \uGN~(NM graphs), we clearly see the benefit brought by leveraging scene graphs. Specifically, even without the image features (\ie, input = $q$, $c$), \uGN~(NM graphs) and \uGN~(VG graphs) can already be on par or even surpass \uGN~(No graphs) with input image features, suggesting that scene graphs, even automatically generated by machines, indeed encode essential visual information for Visual QA.

By further analyzing \uGN~(VG graphs) and \uGN~(NM graphs), we found that including node attributes always improve the performance, especially when no image features are used. We suggest that future scene graph generation should also predict node attributes.

We observe similar trends when applying the \fGN{} model.

\paragraph{GNs without edges.}
We study the case where edges are removed and nodes are isolated: no edge updates are performed; node and global updates do not consider edge features. The result with \fGN~(VG graphs) is $61.3\%$, compared to $62.5\%$ in Table~\ref{tab:my_label}, justifying the need to take the relationships between nodes into account for Visual QA. 

\paragraph{Stacked GNs.}
We investigate stacking multiple GN blocks for performance improvement. We stack two GN blocks: the updated edges and nodes of the first GN block are served as the inputs to the second one. The intuition is that the messages will pass through the scene graph with one more step to support complicated reasoning. When paired with stacked GNs, \fGN~(VG graphs) achieves a gain of $+0.4\%$ in overall accuracy compared to $62.5\%$.

\begin{table}[]
    \centering
    \small
    \begin{tabular}{l r c}
    \multicolumn{3}{c}{\textbf{(a) Unfactorized models}}\\
    \hline
      Methods & Input &  Accuracy \\
      \hline
      MLP~\cite{chao2017being}  & i, q, c & 58.5 \\
      HieCoAtt~\cite{chao2017being}  & i, q, c & 57.5 \\
      Attntion~\cite{chao2017being}  & i, q, c & 60.1 \\
      \hline
      \multicolumn{3}{c}{\uGN{} (No graphs)}\\
      \hline
        -     & q, c & 43.3 \\
        -    & i, q, c  & 58.3 \\
      \hline
      \multicolumn{3}{c}{\uGN{} (NM graphs)}\\
      \hline
       Name  & q, c & 57.9\\
       Name  & i, q, c & 60.5\\
      \hline
      \multicolumn{3}{c}{\uGN{} (VG graphs)}\\
      \hline
        Name        & q, c  & 60.5 \\
        Name         & i, q, c & 61.9 \\
        Name + Attr  & q, c  & 62.2 \\
        Name + Attr    & i, q, c & 62.6 \\
      \hline
    \end{tabular}
    \hfill
    \begin{tabular}{l r c}
    \multicolumn{3}{c}{\textbf{(b) Factorized models}}\\
    \hline
      Methods & Input & Accuracy \\
      \hline
      fPMC(MLP)~\cite{hu2018learning}  & i, q, c  & 57.7 \\
      fPMC(SAN)~\cite{hu2018learning}  & i, q, c  & 62.6 \\
      fPMC(SAN$\star$)~\cite{hu2018learning}  &  i, q, c  & 63.4 \\
       \hline
      \multicolumn{3}{c}{\fGN{} (No graphs)}\\
      \hline   
      -  & q, c   & 44.8 \\
      -  & i, q, c  & 59.4 \\
      \hline
      \multicolumn{3}{c}{\fGN{} (NM graphs)}\\
      \hline
      Name     & q, c  & 57.6 \\
      Name & i, q, c & 60.0 \\
      \hline
      \multicolumn{3}{c}{\fGN{} (VG graphs)}\\
      \hline
      Name        & q, c  & 60.1 \\
      Name           & i, q, c & 60.7 \\
      Name + Attr       & q, c  & 61.9 \\
      Name + Attr      & i, q, c & 62.5 \\
      \hline
    \end{tabular}
    \caption{\small  Visual QA accuracy (\%) on qaVG with unfactorized and factorized models. Input: global features. NM: neural motifs~\cite{zellers2018neural} graphs. For \uGN{} and \fGN, we always consider the relationship names on edges except for the no graph case. We note that fPMC(SAN) uses LSTM~\cite{hochreiter1997long} to encode questions and fPMC(SAN$\star$) additionally uses it to encode answers.}
    \label{tab:my_label}
    \vskip -10pt
\end{table}

\begin{table}
\tabcolsep 2pt
    \centering
    \small
        \begin{tabular}{ l l r| c c c c c c c c c}
            \multicolumn{3}{c|}{Question type} & What  & Color & Where & Number & How & Who & When & Why & Overall \\ \hline
            \multicolumn{3}{c|}{Percentage} & (46\%) & (14\%) & (17\%) & (8\%) & (3\%) & (5\%) & (4\%) & (3\%) & (100\%) \\
            \hline
            \multicolumn{12}{c}{\uGN} \\
            \hline
            NG  &+& (q, c)     & 40.3 & 50.6 & 36.2 & 52.0 & 41.1 & 37.6 & 83.2 & 39.5 & 43.3 \\
            NG  &+& (i, q, c)    & 57.8 & 59.5 & 59.1 & 55.5 & 45.4 & 56.6 & 84.6 & 48.3 & 58.3 \\
            \hline
            NM(N)  &+& (i, q, c)   & 59.4 & 58.2 & 60.3 & 63.4 & 54.3 & 66.6 & 85.3 & 48.1 & 60.5 \\
            \hline
            VG(N) &+& (q, c)      & 61.6 & 54.0 & 62.4 & 58.6 & 45.9 & 63.9 & 83.2 & 50.3 & 60.5 \\
            VG(N)  &+& (i, q, c)   & 61.1 &	61.4 & 62.3	& 59.4 & 54.3 &	67.5 & 85.3	& 48.9 & 61.9 \\
            VG(N, A)  &+& (i, q, c) & 61.4 &	63.8 & 62.6 & 61.5 & 54.8 & 67.5 & 84.8 & 49.6 & 62.6 \\
            \hline
            
            \multicolumn{12}{c}{\fGN} \\
            \hline
            NG  &+& (q, c)     & 41.4 & 52.6 & 38.7 & 53.4 & 42.2 &	39.2 & 83.4 & 40.2 & 44.8 \\
            NG  &+& (i, q, c)    & 58.7 &	61.0 & 60.4 & 57.4 & 47.1 &	57.6 & 85.8 & 49.8 & 59.4 \\
            \hline
            NM(N)  &+& (i, q, c)  & 58.7 &	60.8 & 60.4 & 60.1 & 47.2 &	61.8 & 84.8 & 49.0 & 60.0 \\
            \hline
            VG(N) &+& (q, c)    & 60.9 & 53.6 & 62.0 & 58.1 & 46.2 & 63.3 & 83.7 & 50.9 & 60.1 \\
            VG(N)  &+& (i, q, c)   & 60.2 & 60.4 & 61.8 & 58.5 & 47.4 &	63.8 & 85.1 & 49.6 & 60.7 \\
            VG(N, A)  &+& (i, q, c) & 61.0 &	64.4 & 62.4 & 58.8 & 48.2 &	64.2 & 85.6 & 51.2 & 62.5 \\
            \hline
        \end{tabular}
    \caption{\small Visual QA accuracy (\%) on different question types. VG: Visual Genome~\cite{krishna2017visual} graphs. NM: neural motifs~\cite{zellers2018neural} graphs. NG: no graphs. N: node names. A: node attributes.}
    \label{qType}
    \vskip -10 pt
\end{table}

\subsection{Analysis}\label{sec:resAnalysis}
We provide detailed results on \emph{qaVG} with different question types in Table~\ref{qType}. We found that without input image features, scene graphs with node names significantly improve all question types but ``when'', which needs holistic visual features. Even with image features, scene graphs with node names can still largely improve the ``what'', ``who'', and ``number'' types; the former two take advantage of node names while the later takes advantage of number of nodes.
Adding the node attributes specifically benefits the ``color'' type.
Overall, the VG graphs are of higher quality than NM graphs except for the ``number'' type. We surmise that well-trained object detectors adopted in~\cite{zellers2018neural} can capture smaller objects easily while human annotators might focus on salient ones. 

\begin{figure*}
    \centerline{\includegraphics[width=1\linewidth]{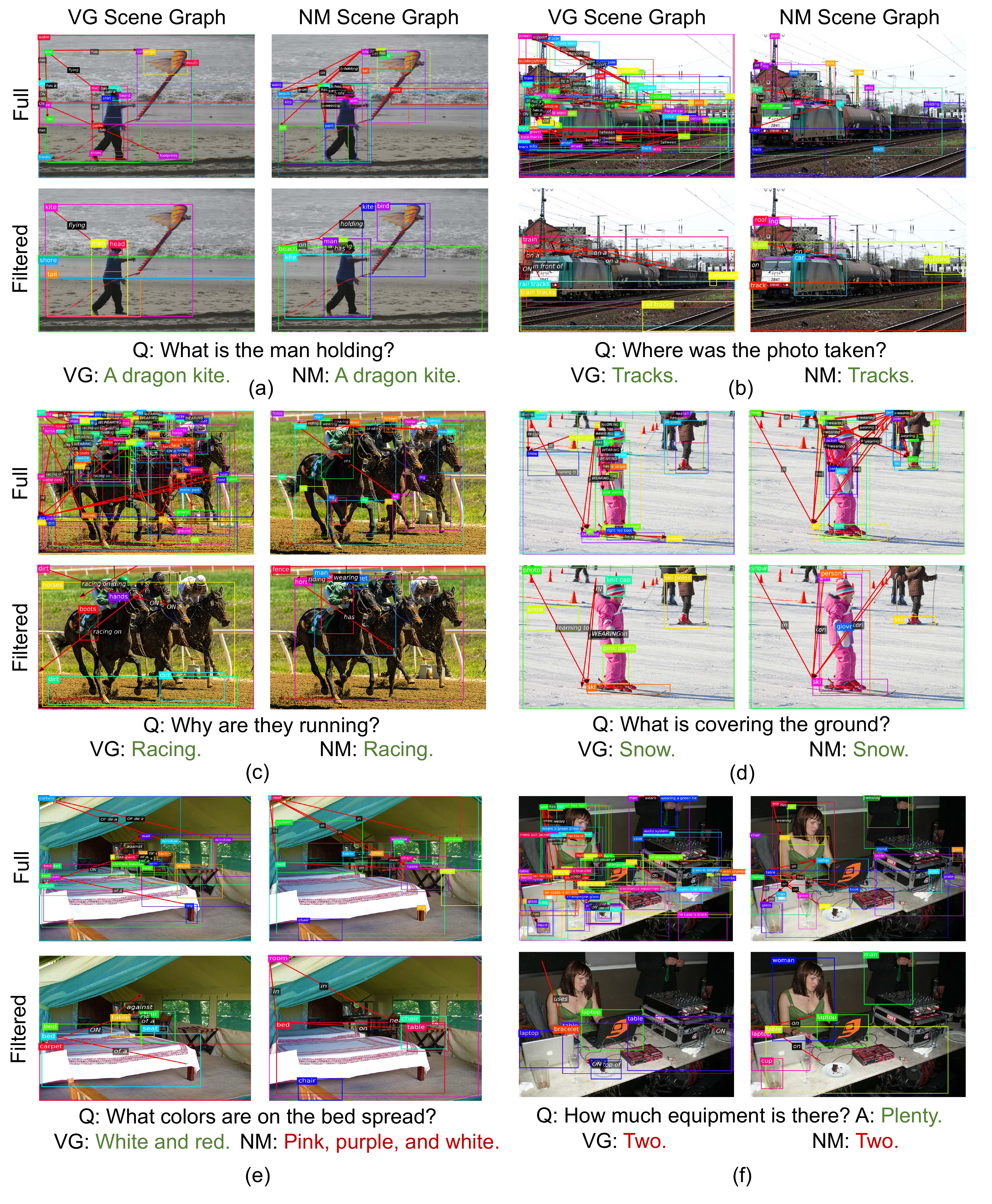}}
    \caption{\small Qualitative results (better viewed in color). We show the original scene graphs (full) and the filtered ones by removing updated nodes and edges with smaller $\ell_2$ norms. Correct answers are in green and incorrect predictions are in red. VG: Visual Genome~\cite{krishna2017visual} graphs. NM: neural motifs~\cite{zellers2018neural} graphs. GN-based Visual QA models can implicitly attend to nodes and edges that are related to the questions (\eg, \textit{kite} and \textit{holding} in (a), \textit{tracks} in (b), \textit{racing} and \textit{riding} in (c), and \textit{snow} and \textit{ground} in (d)). The failure case with NM graphs in (e) is likely due to that no node attributes are provided. The challenge of Visual QA in (f) is the visual common sense: how to connect a coarse-grained term (\eg, \textit{equipment}) with fine-grained terms (\eg, \textit{laptop}, \textit{audio device} and so on). Zoom in for details.}
    \label{f3}
    \vskip -10 pt
\end{figure*}

\paragraph{Qualitative results.}
Fig.~\ref{f3} shows several qualitative examples. Specifically, we only show nodes and edges that have higher $\ell_2$ norms after updates. We see that GN-based Visual QA models (\uGN) can implicitly attend to nodes and edges related to the questions, revealing the underlying reasoning process. 

\section{Conclusion}
\label{sec:disc}
 
In this paper we investigate if scene graphs can facilitate Visual QA.
We apply the graph network (GN) that can naturally encode information on graphs and perform structured reasoning. Our experimental results demonstrate that scene graphs, even automatically generated by machines, can definitively benefit Visual QA if paired with appropriate models like GNs. Specifically, leveraging scene graphs largely increases the Visual QA accuracy on questions related to counting, object presence and attributes, and multi-object relationships. We expect that the GN-based model can be further improved by incorporating image features on nodes as well as advanced multi-modal fusion and attention mechanisms.

\paragraph{Acknowledgements.}
The computational resources are supported by the Ohio Supercomputer Center (PAS1510) ~\cite{OhioSupercomputerCenter1987}.

{\footnotesize
\bibliographystyle{bmvc2k}
\bibliography{main}
}

\clearpage

\appendix

\section*{Supplementary Material}

In this Supplementary Material, we provide details omitted in the main text.
\begin{itemize}
	\item Section~\ref{sec:imple}: Implementation details (Sect.~\ref{sec:setup} of the main text).
	\item Section~\ref{sec:res}: Additional experimental results (Sect.~\ref{sec:mainRes} and Sect.~\ref{sec:resAnalysis} of the main text).
\end{itemize}

\section{Implementation Details}
\label{sec:imple}
In this section, we provide more details about the configuration of scene graph generation, scene graph encoding, the stacked GN model, and the corresponding training procedures.

\subsection{Configuration of scene graph generation}
\paragraph{Human annotated scene graph.}
We leverage the ground truth scene labels of the Visual Genome (VG) dataset~\cite{krishna2017visual} as our human annotated scene graphs\footnote{The VG scene graphs are obtained from \url{https://visualgenome.org/}without any modifications.}.

\paragraph{Machine generated scene graph.}
The commonly used data split of scene graph generation research is different from $qaVG$~\cite{chao2017being}\footnote{The $qaVG$ follows the same data split of Visual Genome dataset for Visual QA task.}. Thus we retrain the start-of-the-art Neural Motifs (NM)~\cite{zellers2018neural} model only using the training images from $qaVG$. Specifically, we retrain both object detector and relationship classifier of the NM to ensure the model never uses testing images for training. We run the well-trained NM model on $qaVG$ to obtain the machine generated scene graph by removing entities (<0.2) and relationships (<0.1) with small probabilities. We show the scene graph detection performance in Table~\ref{tab:nm}.

\begin{table}[h]
\tabcolsep 2pt
    \centering
    \small
        \begin{tabular}{ l | c c c }
             \centering{Models} & R@20  & R@50 & R@100 \\
            \hline
            Released~\cite{zellers2018neural}     & 21.7 & 27.3 & 30.5 \\
            Retrained~(ours)                      & 21.5 & 27.5 & 30.6 \\
            \hline
        \end{tabular}
    \caption{\small Scene graph detection accuracy (\%) using recall@K metrics. The released NM model~\cite{zellers2018neural} is evaluated on the split of~\cite{xu2017scene}. Our retrained NM model is evaluated on the split of $qaVG$~\cite{chao2017being}. We can see that the machine generated scene graph from the retrained NM model achieves satisfied performance.}
    \label{tab:nm}
\end{table}

\subsection{Scene graph encoding}
As mentioned in Sect. 4.1 of the main text, we extract corresponding features to represent nodes, edges, and global. To represent image $i$, we extract the activations from the penultimate layer of the ResNet-152~\cite{he2016deep} pretrained on ImageNet~\cite{russakovsky2015imagenet} and average them to obtain a 2,048-dimentional feature representation. The question $q$, candidate $c$, node names, node attributes, and edge names are represented as the average word to vector~\cite{mikolov2013distributed} embeddings. Specifically, we remove punctuation, change all characters to lowercases, and convert all integer numbers within [0, 10] to words before computing word to vector. We use `UNK' to represent out-of-vocabulary (OOV) word. Finally, the NM scene graph can be represented as 300-dimentional nodes and 300-dimentional edges embeddings, and the VG scene graph has 600-dimentional nodes and 300-dimentional edges representations. To enable better generalization on unseen datasets, we fix all the visual and language features in our experiments. All individual features are $\ell_2$ normalized before concatenation.

\subsection{Stacked GNs}
We provide more details about stacked GNs. As mentioned in Sect. 4.2 of the main paper, the overall accuracy could be improved with respect to the number of GN blocks. Here, we propose one design choice of the stacked GNs as below,
\begin{align}\label{stacked-gn}
    & (u, V', E') = GN_{f^{e}_\theta,f^{v}_\theta}(u, V, E), \\ 
    & (u', V'', E'') = GN_{f^{e}_\theta,f^{v}_\theta,f^{u}_\theta}(u, V',E'),\nonumber
\end{align}
where the first GN block only performs edge and node updating, and the updated properties will be served as the inputs of the second one. Finally, the resulting global feature $u'$ can be used to perform inference in Visual QA. Similarly, multiple GN blocks can be stacked in such manner. We expect varied designs of multi-layer GN could be proposed for performance improvement, such as jointly learning global feature within the latent GN blocks and making GN blocks recurrent.

\subsection{Optimization}
For all above models, we train for at most 30 epochs using stochastic gradient optimization with Adam~\cite{kingma2014adam}. The initial learning rate is $10^{-3}$, which is divided by 10 after $M$ epochs. We turn the $M$ on the validation set and choose the best model via validation performance.

Within each mini-batch, we sample 100 $(i, q, a)$ triplets. 
In order to prevent unbalanced training\footnote{In $qaVG$ ~\cite{chao2017being}, each $(i, q)$ pair contains 3 QoU-decoys (incorrect answers), 3 IoU-decoys (incorrect answers), and 1 target (correct answer). The machine tends to predict the dominant label if the training is performed among all samples.}, we follow the sampling strategy suggested by Chao~\etal~\cite{chao2017being}. We randomly choose to use QoU-decoys or IoU-decoys for each triplet as negative samples when training. Then the binary classifier is trained on top of the target and 3 decoys for each triplet. That is, 100 triplets in the each mini-batch related to 400 samples with binary labels. In the testing stage, we evaluate the performance of picking the correct answer from all 7 candidates.

\section{Additional Results}
\label{sec:res}
\begin{figure*}[b]
    \centerline{\includegraphics[width=0.8\linewidth]{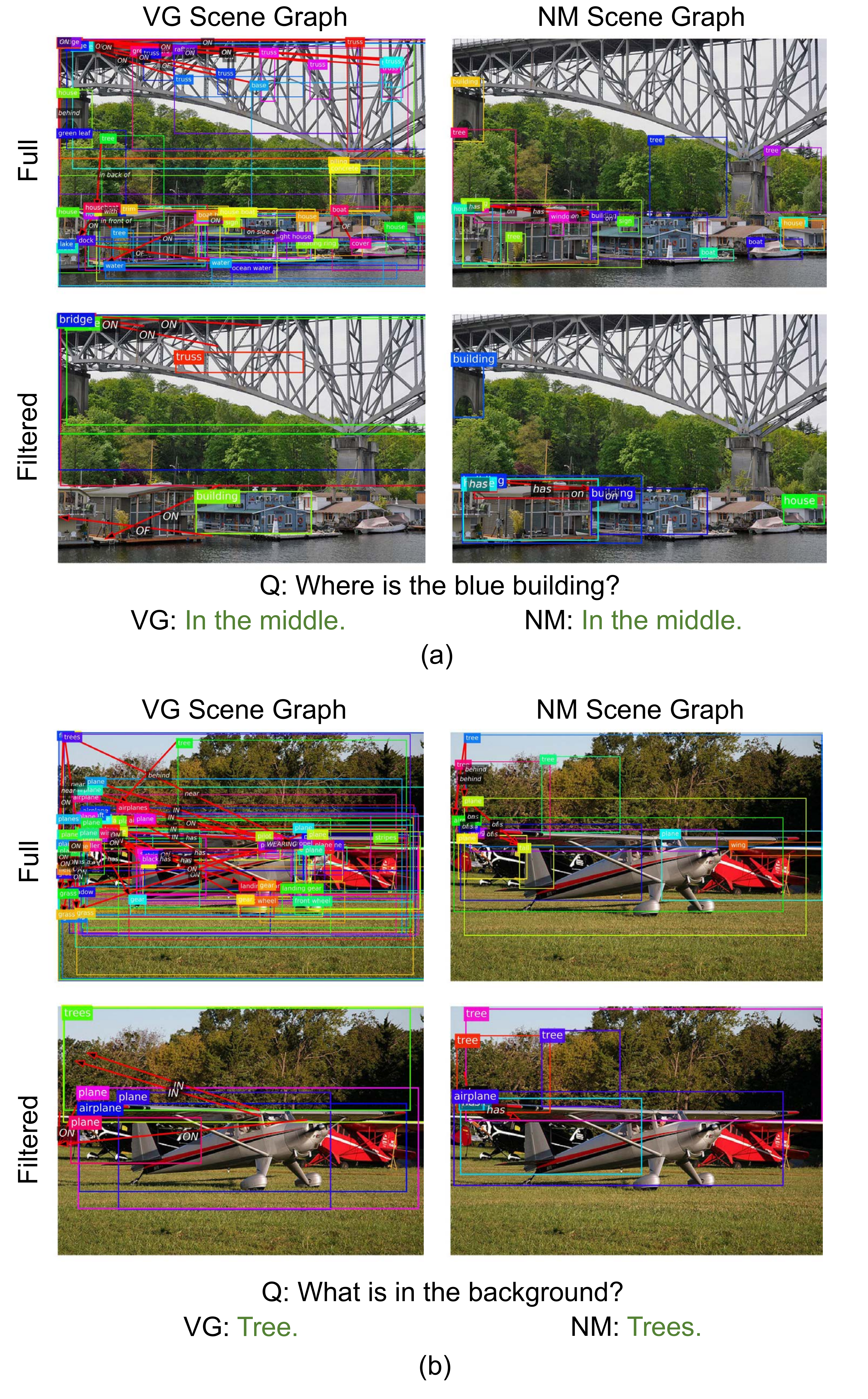}}
    \caption{Qualitative results. (a) Both VG and NM graphs can attend to the \textit{building in the middle}. (b) The models can attend to the \textit{trees} behind the foreground objects.}
    \label{sp1}
\end{figure*}

\begin{figure*}[b]
    \centerline{\includegraphics[width=0.8\linewidth]{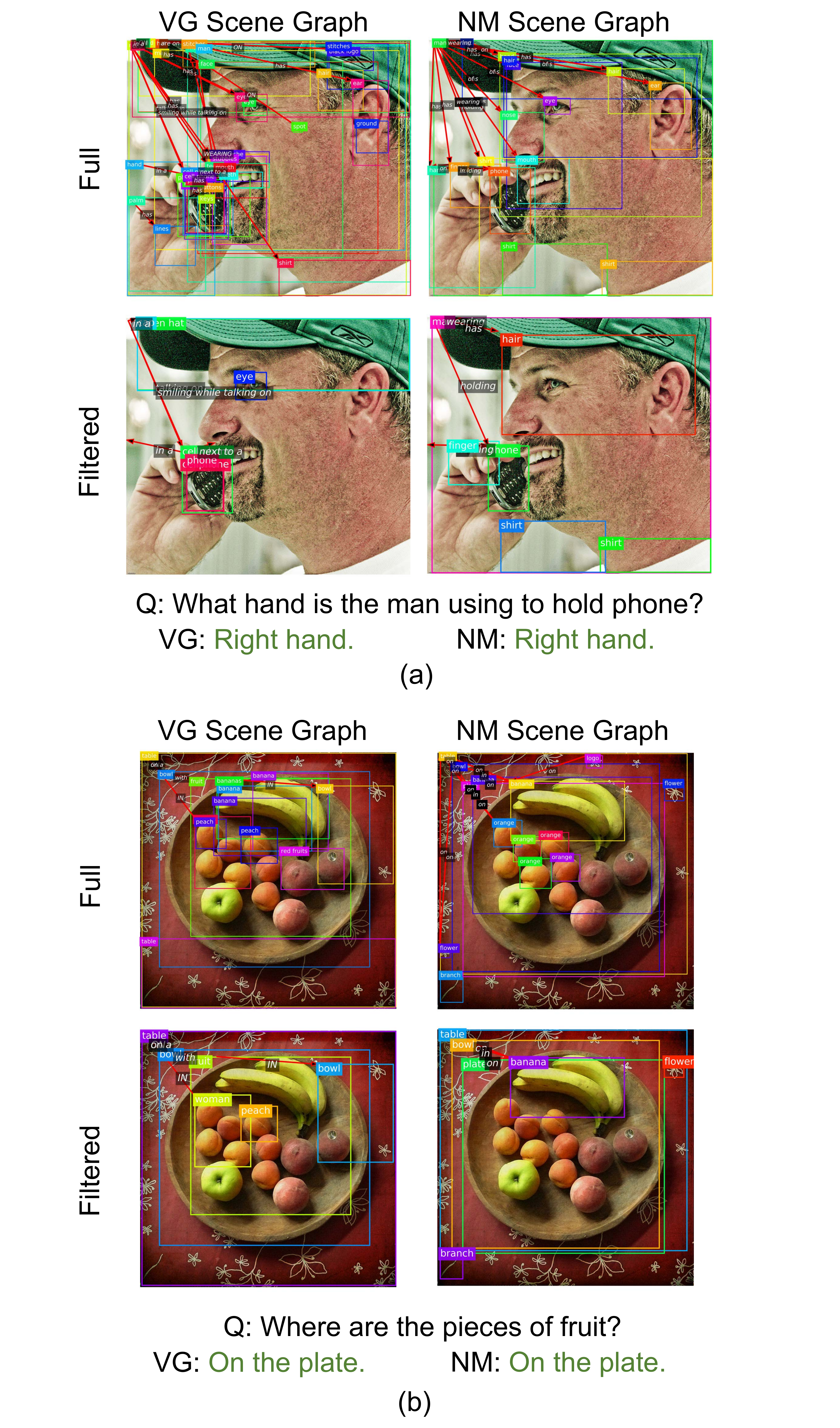}}
    \caption{Qualitative results. (a) Both two scene graphs show the location and relationships of the phone. (b) Both VG and NM graphs attend to the location of the fruits.}
    \label{sp2}
\end{figure*}

\begin{figure*}[b]
    \centerline{\includegraphics[width=0.8\linewidth]{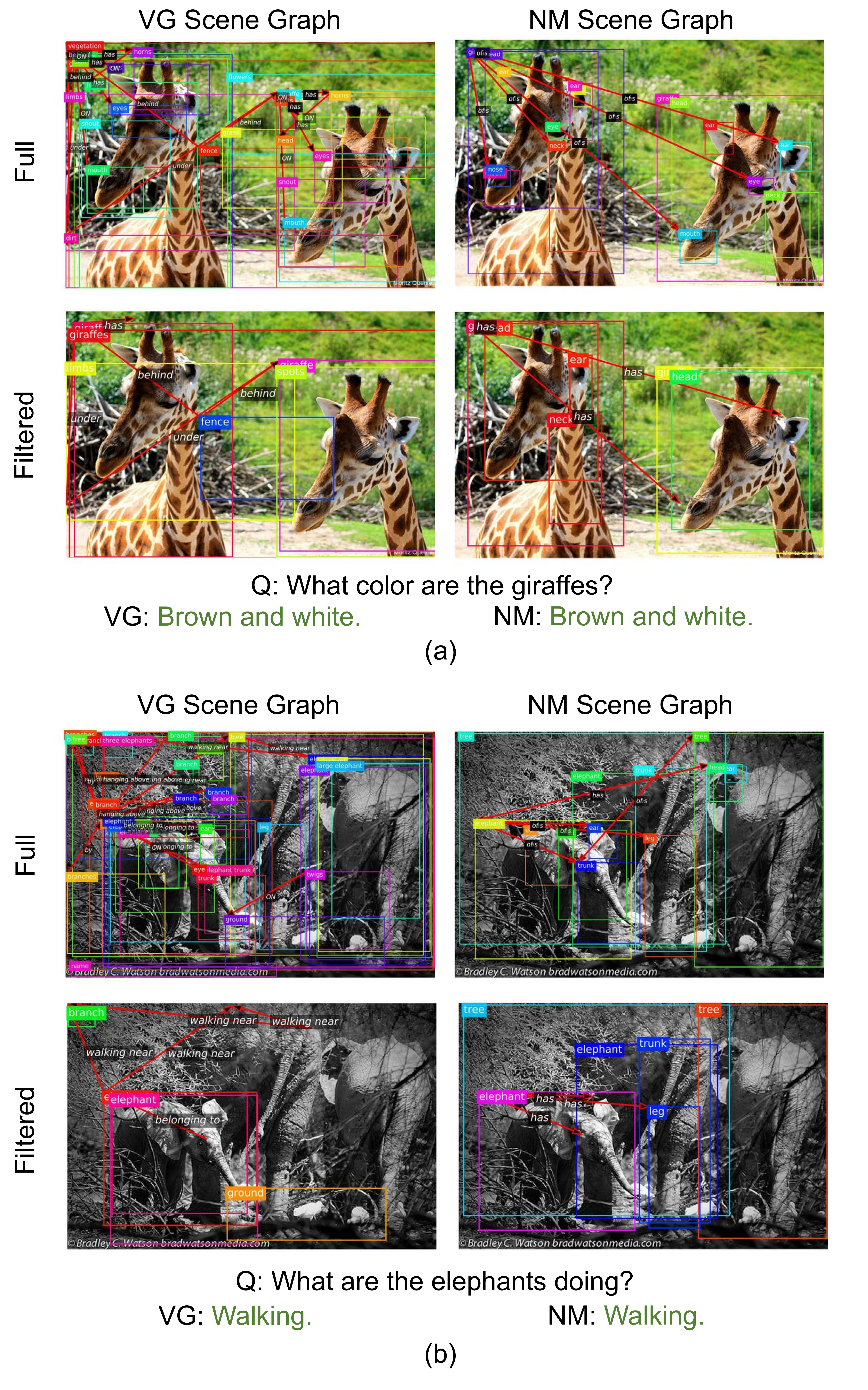}}
    \caption{Qualitative results. (a) Both VG and NM graphs attend to the giraffes. Even without color attributes, the NM scene graph may learn the colors through overall image features. (b) VG graph clearly captures \textit{walking near} and NM graph attends to \textit{elephants}.}
    \label{sp3}
\end{figure*}

\begin{figure*}[b]
    \centerline{\includegraphics[width=0.7\linewidth]{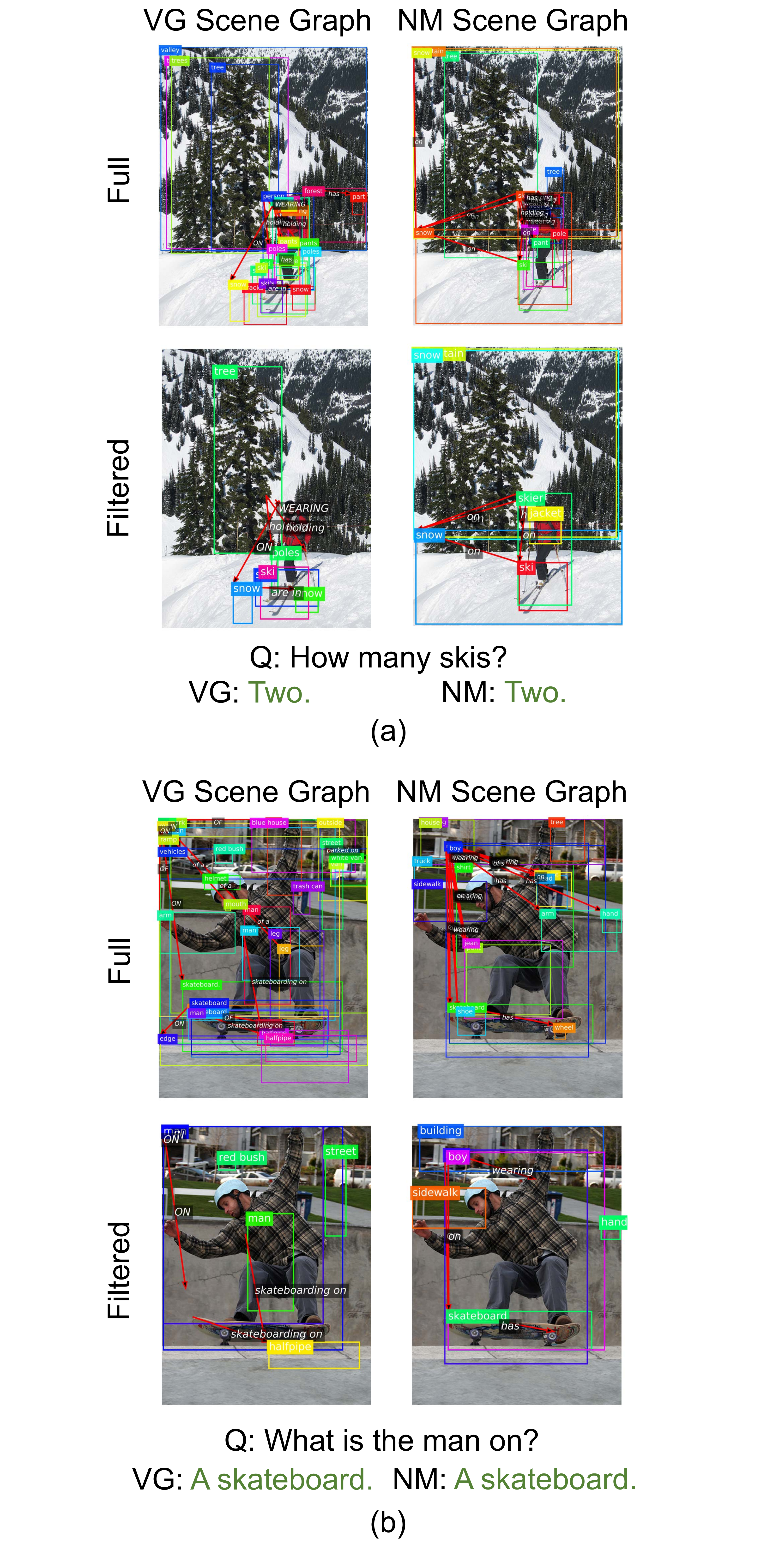}}
    \caption{Qualitative results. (a) Both two graphs capture \textit{two skis} from the full scene graphs. (b) VG graph attends to the relationship \textit{skateboarding on} and NM graph attends to \textit{boy on skateboard}.}
    \label{sp4}
\end{figure*}

\begin{figure*}[b]
    \centerline{\includegraphics[width=0.9\linewidth]{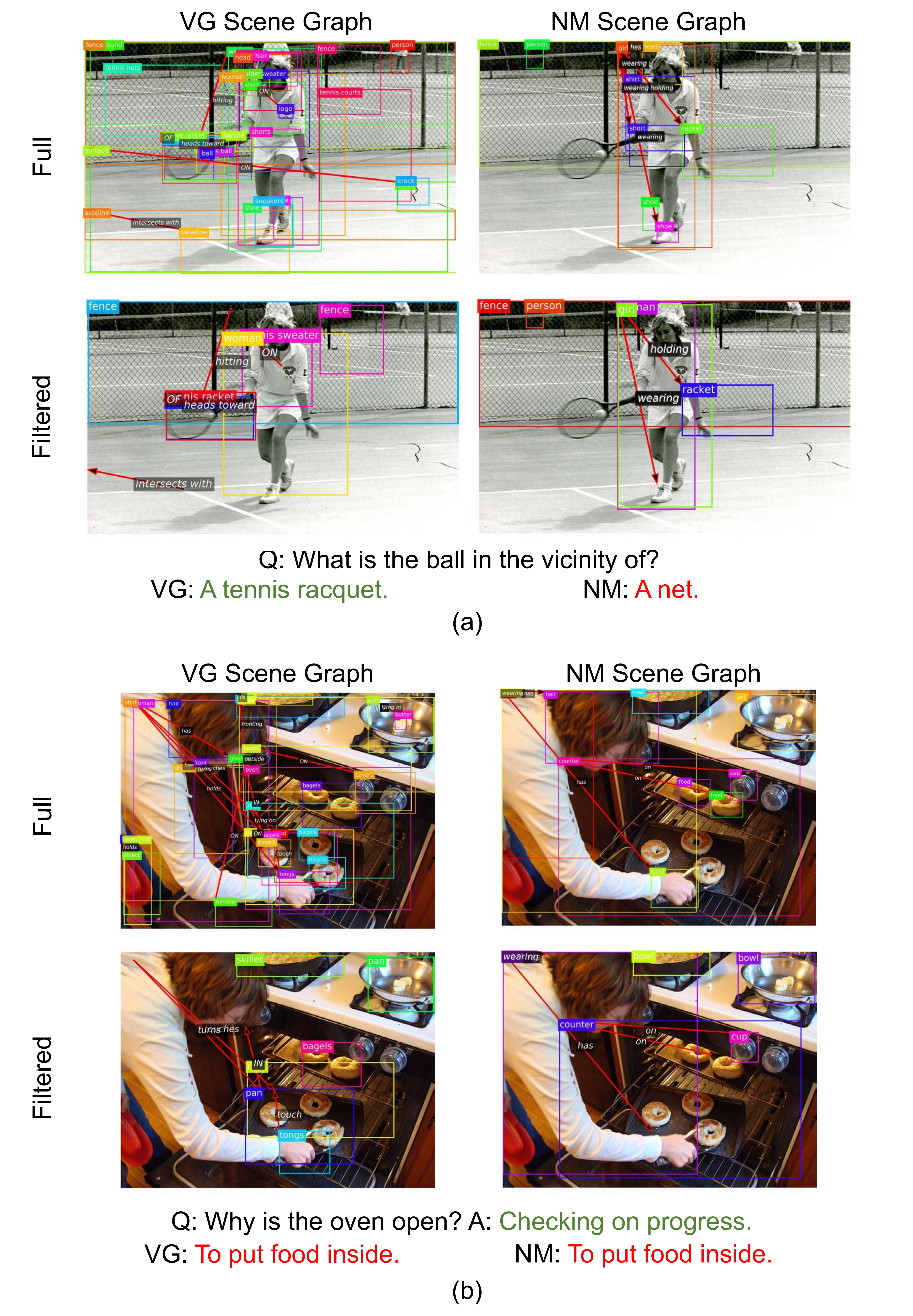}}
    \caption{Failure cases. (a) VG scene graph attends to \textit{heads toward} and \textit{tennis racket} to predict the correct answer while NM graph fails because of lacking such high level semantic annotations. (b) The model needs to understand the interaction between the person and the food inside the oven, which is a hard case.}
    \label{sp5}
\end{figure*}

In this section, we provide more qualitative results. In Fig.~\ref{sp1},~\ref{sp2},~\ref{sp3},~\ref{sp4},~\ref{sp5}, we show the original scene graphs (full) and the filtered ones by removing updated nodes and edges with smaller $\ell_2$ norms. Correct answers are in green and incorrect predictions are in red. VG: Visual Genome~\cite{krishna2017visual} graphs. NM: neural motifs~\cite{zellers2018neural} graphs. 

\end{document}